\documentclass[letterpaper, 10 pt, conference]{ieeeconf}  %

\IEEEoverridecommandlockouts                              %

\usepackage{hyperref}

\usepackage{svg} %
\usepackage{graphicx}
\usepackage{graphics} %
\usepackage{bbm}
\usepackage{mwe}%
\usepackage{float}
\usepackage{amsmath} %
\usepackage{amssymb}  %
\usepackage{color}
\usepackage{xspace}
\usepackage{booktabs}
\usepackage{multirow} 
\usepackage{graphicx,subcaption,lipsum}
\usepackage[
    style=ieee,
    natbib=true,
    citestyle=numeric-comp,
    doi=false,
    isbn=false,
    url=false]{biblatex} 

\usepackage{algorithm}
\usepackage{algorithmic}
\usepackage{cleveref}
\usepackage{wrapfig}

\usepackage{siunitx}
\usepackage[inter-unit-product =\cdot]{siunitx}
\usepackage{threeparttable}
\usepackage{booktabs, caption, makecell}

\newif\ifincludenote
\includenotetrue
\ifincludenote
    \usepackage{marginnote}
    \newcommand{\cynote}[1]{\textcolor{violet}{(Chaoyi: #1)}}
    
\else
    \newcommand{\cynote}[1]{}
\fi

\makeatletter
\let\NAT@parse\undefined
\makeatother

\definecolor{mydarkblue}{rgb}{0,0.08,0.45}
\definecolor{mydarkgreen}{RGB}{0, 139, 69}
\definecolor{mygreen2}{RGB}{0 205 0}
\definecolor{mybrown}{RGB}{139 69 19}
\hypersetup{
	colorlinks=true,
	linkcolor=blue,
	urlcolor=magenta,
	citecolor=mygreen2,
}

\newcommand{\method}{DIAL-MPC\xspace}

\newtheorem{proposition}{Proposition}

\Crefname{asm}{Assumption}{Assumption}

\title{\LARGE \bf
Full-Order Sampling-Based MPC for Torque-Level \\ Locomotion Control via Diffusion-Style Annealing
}

\author{}
\author{Haoru Xue$^{*}\thanks{* These authors contributed equally to this work.}$, Chaoyi Pan$^{*}$, Zeji Yi, Guannan Qu, and Guanya Shi 
\thanks{The authors are with Carnegie Mellon University, USA. {\tt\small\{haorux, chaoyip, zejiy, gqu, guanyas\}@andrew.cmu.edu}.}
\thanks{Paper website: \href{https://lecar-lab.github.io/dial-mpc/}{https://lecar-lab.github.io/dial-mpc/}}
}

\begin{document}

\maketitle

\thispagestyle{empty}
\pagestyle{empty}

\begin{abstract}

Due to high dimensionality and non-convexity, real-time optimal control using full-order dynamics models for legged robots is challenging. 
Therefore, Nonlinear Model Predictive Control (NMPC) approaches are often limited to reduced-order models.
Sampling-based MPC has shown potential in nonconvex even discontinuous problems, but often yields suboptimal solutions with high variance, which limits its applications in high-dimensional locomotion.
This work introduces DIAL-MPC (Diffusion-Inspired Annealing for Legged MPC), a sampling-based MPC framework with a novel diffusion-style annealing process. 
Such an annealing process is supported by the theoretical landscape analysis of Model Predictive Path Integral Control (MPPI) and the connection between MPPI and single-step diffusion.
Algorithmically, DIAL-MPC iteratively refines solutions online and achieves both global coverage and local convergence.
In quadrupedal torque-level control tasks, DIAL-MPC reduces the tracking error of standard MPPI by $13.4$ times and outperforms reinforcement learning (RL) policies by $50\%$ in challenging climbing tasks \emph{without any training}. 
In particular, DIAL-MPC enables precise real-world quadrupedal jumping with payload.
To the best of our knowledge, DIAL-MPC is the first \emph{training-free} method that optimizes over full-order quadruped dynamics in real-time.

\end{abstract}

\section{INTRODUCTION}

Legged robots have demonstrated great potential in navigating through complex environments thanks to their agility and mobility~\cite{parkHighspeedBoundingMIT2017,kimHighlyDynamicQuadruped2019,herdtOnlineWalkingMotion2010,khazoomTailoringSolutionAccuracy2024,koenemannWholebodyModelpredictiveControl2015,neunertWholeBodyNonlinearModel2018}.
However, the online control of articulated legged systems remains challenging because of their high-dimensional, underactuated and contact-rich nature, leads to non-convex and non-smooth optimization landscapes.

\begin{figure}[h]
    \centering
    \includegraphics[width=\linewidth]{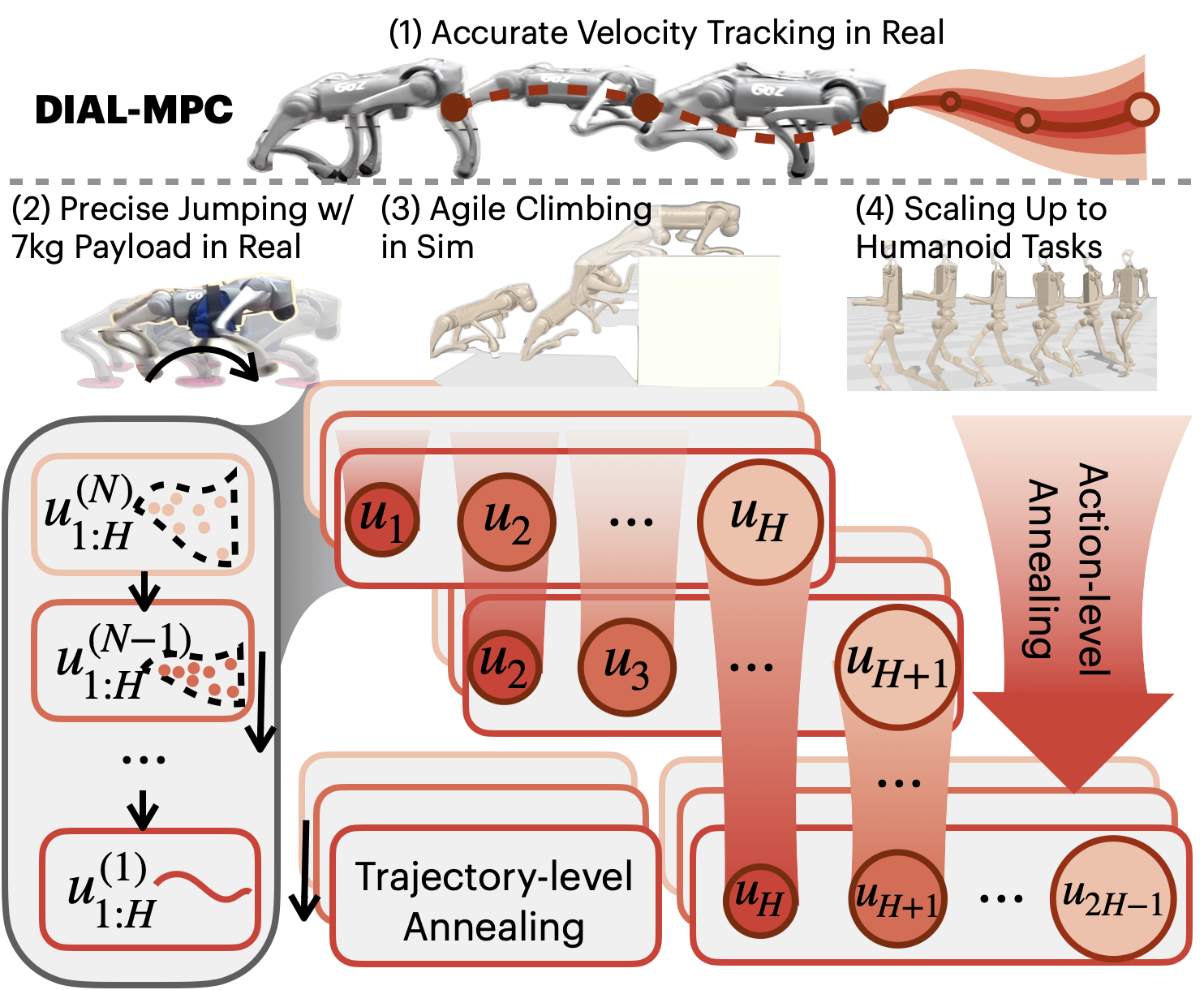}
    \caption{
    Diffusion-inspired annealing for legged MPC (\method).
    To achieve both global coverage and local convergence, \method involves a bi-level diffusion-inspired annealing process. Trajectory-wise annealing is performed with different sampling variance. Action-wise annealing is performed on control input at different horizion.
    Over time, $u_H$ will be gradually refined by the two diffusion-inspired annealing processes, leading to a robust and efficient full-order online control.
    }
    \label{fig:sync_async}
    \vspace{-0.5cm}
\end{figure}

Reinforcement learning (RL) has become a popular approach for learning control policies for legged robots, thanks to its ease of implementation and strong performance in contact-rich problems~\cite{radosavovicHumanoidLocomotionNext2024,heAgileSafeLearning,chengExtremeParkourLegged2024,mikiLearningRobustPerceptive2022,rudinLearningWalkMinutes,chenLearningTorqueControl2023,fuMinimizingEnergyConsumption2021}.
However, RL suffers from time-consuming training and tedious tuning, and the resulting policies heavily depend on the training setup, limiting their test-time generalization to unseen tasks and environments.
In the meanwhile, NMPC~\cite{neunertWholeBodyNonlinearModel2018,kuindersmaEfficientlySolvableQuadratic2014,koenemannWholebodyModelpredictiveControl2015} is often limited to reduced-order models due to the 
intractability of solving full-order problems involving contacts and nonlinear dynamics. 

Sampling-based MPC~\cite{williamsAggressiveDrivingModel2016,yiCoVOMPCTheoreticalAnalysis2024,yinTrajectoryDistributionControl2022,williamsRobustSamplingBased2018} has been applied to various nonlinear and hybrid dynamical systems because of its flexibility in handling arbitrary dynamics and constraints, as well as parallelizability. 
Nonetheless, these algorithms are sensitive to hyperparameters in high-dimensional and non-convex optimization problems, particularly the sampling kernel, leading to high variance and suboptimal performance. 
Specifically, a large sampling range provides better global coverage but may result in solutions far from the optimum, while a small sampling range enhances local search ability but is more susceptible to local minima and initial guesses, leading to increased variance.

In this paper, we address these challenges by unveiling the intrinsic connection between sampling-based MPC and diffusion processes. 
Following the key iterative refinement idea of the diffusion model,
we introduce a novel sampling-based MPC method, \underline{D}iffusion-\underline{I}nspired \underline{A}nnealing for \underline{L}egged \underline{M}odel \underline{P}redictive \underline{C}ontrol (\method).  
\method optimizes control sequences iteratively in a dual-loop manner. 
Specifically, \method starts optimizing the control sequence with smooth but inaccurate objectives and gradually shifts to more accurate local objectives.

Compared with MPPI, \method enables on-the-fly, full-order torque-level locomotion control by effectively balancing global coverage and local convergence. 
We evaluate \method on a quadrupedal robot with full-order dynamics and show that it can achieve real-time 50Hz control with both robustness and efficiency. As a training-free and purely online method, \method enables precise jumping with payload and outperforms RL methods.
The contribution of this paper is three-fold:
\begin{itemize}
    \item \textbf{Novel Diffusion-Inspired Annealing Framework}: We propose a diffusion-inspired annealing framework for sampling-based MPC by revealing the connection between sampling-based MPC and diffusion processes. 
    \item \textbf{Full-Order Torque-Level Control of Legged Robot}: We develop and implement DIAL-MPC for full-order torque-level control of legged robots based on the proposed annealing framework. To our knowledge, this is the first framework achieving both real-time flexibility and RL-level agility in legged locomotion.
    \item \textbf{Real-World Validation}: We validate the performance of \method for quadruped control, showing that it achieves real-time 50 Hz control with robustness and efficiency. 
\end{itemize}

\section{RELATED WORK}

\subsection{Agility in Legged Locomotion}
Nonlinear Model Predictive Control (NMPC), particularly gradient-based methods, has demonstrated significant success in real-time control of legged locomotion with closed-loop stability~\cite{parkHighspeedBoundingMIT2017, grandiaPerceptiveLocomotionNonlinear2022, neunertWholeBodyNonlinearModel2018, sleimanUnifiedMPCFramework2021}. These approaches typically employ reduced-order models to mitigate the burden of planning over full-order hybrid dynamics, necessitating lower-level whole-body controllers for motion execution~\cite{kuindersmaEfficientlySolvableQuadratic2014, herdtOnlineWalkingMotion2010, kimHighlyDynamicQuadruped2019, koenemannWholebodyModelpredictiveControl2015}.

However, reduced-order models can lead to sub-optimal performance and constraint violations, particularly under high agility demands. For instance, \cite{khazoomTailoringSolutionAccuracy2024} introduces a tailored solver for full-order NMPC online, yet remains constrained by predefined contact sequences, limiting motion agility and robustness. In contrast, our method enables full-order model MPC without redundant constraints, allowing adaptation to real-time feedback and environmental interactions.

Model-free Reinforcement Learning (RL) has also been explored to address full-order control by learning optimal policies directly from high-fidelity simulations~\cite{heOmniH2OUniversalDexterous, fuHumanPlusHumanoidShadowing, zhangWoCoCoLearningWholeBody, heLearningHumanHumanoidRealTime2024, rudinLearningWalkMinutes, fuMinimizingEnergyConsumption2021}. While these approaches eliminate the need for explicit modeling, they suffer from  limited generalization to new tasks and dynamic environments.

Goal-conditioned RL~\cite{ghoshLearningActionableRepresentations2019, atanassovCurriculumBasedReinforcementLearning2024} enhances task-level generalization but still struggles with unseen dynamics and novel task classes. Our approach overcomes these limitations by enabling rapid motion generation through training-free online optimization, thereby combining the agility of RL with the robustness and generalization capabilities of MPC.

\subsection{Sampling-Based Optimization}
Sampling-based or zeroth-order optimization methods, including Bayesian Optimization~\cite{frazierTutorialBayesianOptimization2018, sohl-dicksteinDeepUnsupervisedLearning2015}, the Cross-Entropy Method~\cite{mannorCrossEntropyMethod}, and Evolutionary Algorithms~\cite{wierstraNaturalEvolutionStrategies2008}, are widely utilized for solving non-convex and non-smooth optimization problems. They differ from first-order methods as the gradient information is no longer required. These methods are particularly effective in applications such as hyperparameter tuning~\cite{snoekPracticalBayesianOptimization2012, hernandez-lobatoParallelDistributedThompson2017} and generative modeling~\cite{songDenoisingDiffusionImplicit2020}. 

In the context of real-time robot control, Model Predictive Path Integral (MPPI) control~\cite{williamsAggressiveDrivingModel2016} and its variants~\cite{yiCoVOMPCTheoreticalAnalysis2024, yinTrajectoryDistributionControl2022, williamsRobustSamplingBased2018} have gained popularity for online motion planning ~\cite{pravitraL1AdaptiveMPPIArchitecture2020, sacksDeepModelPredictive2023, howellPredictiveSamplingRealtime2022}, due to their inherent parallelizability and flexibility.  Given a high-stiffness problem like legged locomotion, zeroth-order methods including policy gradient~\cite{suttonPolicyGradientMethods1999} have shown better convergence properties both empirically and theoretically~\cite{suhDifferentiableSimulatorsGive2022} from their smoothing nature. 

Despite their advantages, sampling-based methods are plagued by the curse of dimensionality and high variance, especially under constrained online sampling budgets. This is particularly problematic in high-dimensional, contact-rich environments.

\subsection{Parallel Robot Simulation}
Massively parallelizable simulation environments, such as Isaac Gym~\cite{makoviychukIsaacGymHigh2021}, Brax~\cite{freemanBraxDifferentiablePhysics2021}, and MuJoCo~\cite{MuJoCoPhysicsEngine}, have become essential tools in the development of zeroth-order optimization methods for robot control. These simulators facilitate the rapid generation of data required by sample-hungry algorithms like PPO~\cite{schulmanProximalPolicyOptimization2017}, enabling efficient training of complex policies.

\section{METHOD}

In this section, we present our method by establishing the equivalence between MPPI and a single-stage diffusion process (\cref{subsec:sampling-mpc}). 
The connection then explains why the annealing process in diffusion helps MPPI to optimize over a non-smooth landscape, as discussed in \cref{subsec:annealing}. 
Leveraging this equivalence, we introduce a diffusion-inspired annealing technique for MPPI in \cref{subsec:annealing_mppi}. 

\subsection{Sampling-Based MPC as Single-stage Diffusion}
\label{subsec:sampling-mpc}

\noindent\textbf{Optimal control problem.}
Sampling-based MPC aims to solve the following optimization problem:
\begin{align*}
    \min_{u_{t:t+H}} J(u_{t:t+H}) & = \sum_{h=0}^{H} c(x_{t+h}, u_{t+h}) + c_f(x_{t+H+1}),       \\
    \text{s.t.} \quad x_{t+h+1}   & = f(x_{t+h}, u_{t+h}) \quad \forall h \in \{0, \ldots, H\}, \\
    x_{t:t+H+1}                   & \in \mathcal{X}, \ u_{t:t+H} \in \mathcal{U}
\end{align*}
where $x_{t+h}$ is the state at time $t+h$, $u_{t+h}$ is the control input at time $t+h$, $f$ is the system dynamics, $c$ and $c_f$ are the cost function and terminal cost function, $\mathcal{X}$ and $\mathcal{U}$ are the state and control constraints, respectively.

MPPI estimates the optimal control sequence through the following steps:
First, draw $N_W$ perturbations from a Gaussian distribution $[W]_{i} \sim \mathcal{N}(0, \Sigma_{t:t+H}), i=1,\ldots, N_W$ (which we collectively denote as $[W]_{1:N_W})$. 
Then the cost function $J(u_{t:t+H})$ is evaluated for each sampled control sequence by rolling out the system dynamics and cumulatively summing the cost.
For each perturbed control sequence  $U+[W]_{i}$, where $U=u_{t:t+H}$, evaluate the cost function  $J(U + [W]_i)$  by simulating the system dynamics and accumulating the costs.
Finally, update the control sequence using temperature $\lambda$:
\begin{align}
    U^+ = U + \frac{\sum_{i=1}^{N_W} \exp\left(-\frac{J(U + [W]_i)}{\lambda}\right) [W]_i}{\sum_{j=1}^{N_W} \exp\left(-\frac{J(U + [W]_j)}{\lambda}\right)},
    \label{eq:mppi}
\end{align}

\begin{figure}[h]
    \centering
    \includegraphics[width=\linewidth]{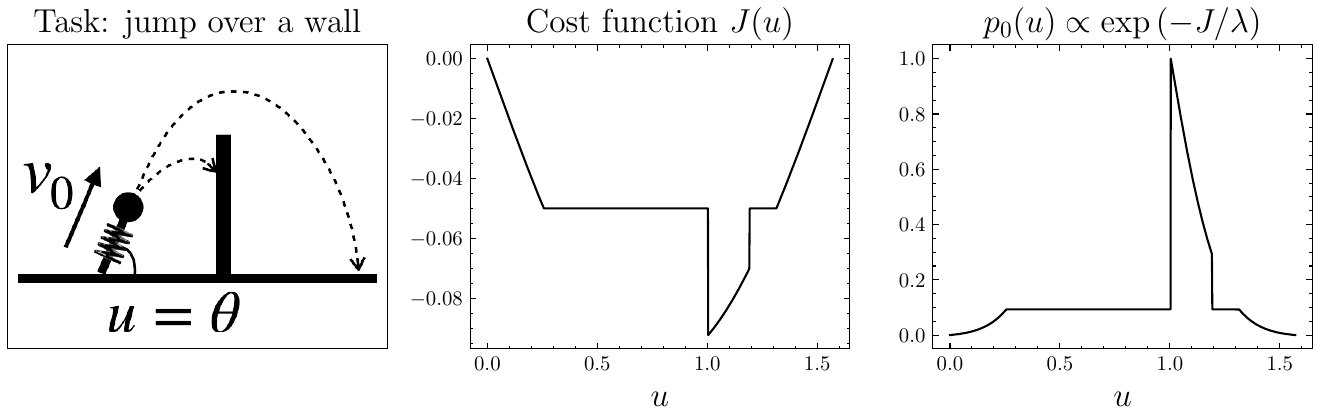}
    \caption{Cost function $J(U)$ and target distribution $p_0(U)$ for a task where robot need to jump over a wall. The cost function could be highly non-convex and non-smooth due to the contact constraint. The resulting distribution $p_0(U)$ is also non-convex and sparse, which is hard to sample from.}
    \label{fig:demo_task}
\end{figure}

\noindent\textbf{MPPI as a single-stage Diffusion.}
The optimization problem can be reframed in a sampling context. Define the target distribution $p_0(U) \propto \exp\left(-\frac{J(U)}{\lambda}\right)$. As $\lambda \to 0$, samples from $p_0(U)$ concentrate around the optimal control sequence $U^*$ as illustrated in \cref{fig:demo_task}.
However, directly sampling from $p_0(U)$ is impractical due to its narrow support. 
To facilitate the sampling, we convolve $p_0(U)$ with a Gaussian noise kernel $\phi(\cdot)$, i.e., the density of $\mathcal{N}(0, \Sigma)$ to get the corrupted distribution $p_1(\cdot) \propto (p_0 \ast \phi)(\cdot) $ as depicted in \cref{fig:demo_compare}.

\begin{proposition}[Adopted from~\cite{panModelBasedDiffusionTrajectory2024}]
    \label{prop:mppi_single_stage}
    The MPPI update \eqref{eq:mppi} can be viewed as a one-step ascent with the score function $\nabla \log p_1(U)$ with a learning rate $\Sigma$:
    \begin{align}
        \label{eq:score_ascent}
        U^+ = U + \Sigma \cdot \nabla \log p_1(U). 
    \end{align}
\end{proposition}

\begin{proof}
    The diffused distribution $p_1$ is defined as $p_1(\cdot) \propto (p_0 \ast \phi)(\cdot) $ as in \cref{fig:demo_compare}. The score function $\nabla \log p_1(U)$ can be calculated in the following way:

    \begin{subequations}\label{eq:score_derivation}\small
        \begin{align}
                    & \nabla \log p_1(U)
            = \frac{\nabla p_1(U)}{p_1(U)}
            = \frac{\nabla \left(p_0(U) \ast \phi(U) \right)}{p_1(U)} \label{eq:score_p1}                                                                                                           \\
            =       & \frac{p_0(U) \ast \nabla \phi(U)}{p_1(U)}
            = -\frac{p_0(U) \ast (\phi(U) \Sigma^{-1}U)}{p_1(U)} \label{eq:score_p2}                                                                                                                \\
            =       & - \Sigma^{-1} \frac{\int p_0(U-W) \phi(W) W dW}{\int p_0(U-W) \phi(W) dW} \label{eq:score_p3}                                                                                 \\
            =       & - \Sigma^{-1} \frac{\mathbb{E}_{W\sim\phi(\cdot)} \left[ p_0(U-W) W \right]}{\mathbb{E}_{W\sim\phi(\cdot)} \left[ p_0(U-W) \right]} \label{eq:score_p4}                       \\
            =       & \Sigma^{-1} \frac{\mathbb{E}_{W\sim\phi(\cdot)} \left[ p_0(U+W) W \right]}{\mathbb{E}_{W\sim\phi(\cdot)} \left[ p_0(U+W) \right]} \label{eq:score_p5}                         \\
            \approx & \Sigma^{-1} \frac{\sum_{i=1}^{N_W} \exp\left(-\frac{J(U + [W]_i)}{\lambda}\right) [W]_i}{\sum_{j=1}^{N_W} \exp\left(-\frac{J(U + [W]_j)}{\lambda}\right)}. \label{eq:score_p6}
        \end{align}
        \label{eq:score}
    \end{subequations}

From \eqref{eq:score_p1} to \eqref{eq:score_p2}, we move the gradient into the convolusion. In \eqref{eq:score_p2}, the gradient of Gaussian kernal $\phi(W)$ is calculated. From \eqref{eq:score_p2} to \eqref{eq:score_p3}, we use the definition of convolution. In \eqref{eq:score_p4}, we rewrite the integral as an expectation.
In \eqref{eq:score_p5}, we flip the sign of $W$ to match the form of MPPI since the Gaussian distribution is symmetric.
From \eqref{eq:score_p5} to \eqref{eq:score_p6}, Monte Carlo approximation is applied.

\end{proof}

In other words, MPPI performs a ``denoising'' step with the score function $\nabla \log p_1(U)$ conditioned on a \emph{fixed} noise level $\mathcal{N}(0,\Sigma)$~\cite{song2019generative}. The key difference is that the score-based diffusion model~\cite{song2019generative} iteratively refines samples using different noise levels, which motivates our annealing design\footnote{There are some other differences between  \eqref{eq:score_ascent} and diffusion models: \eqref{eq:score_ascent} does not have scaling factor or extra noise. See more discussion in~\cite{panModelBasedDiffusionTrajectory2024}.}.

\subsection{Diffusion-Inspired Annealing}
\label{subsec:annealing}

Given that MPPI is a single-stage diffusion that moves particles toward the stationary point of \( p_1(\cdot) = (p_0 \ast \phi)(\cdot) \) (\cref{prop:mppi_single_stage}), a natural question arises: What are the pros and cons of optimizing \( p_1 \) compared to directly solving for the optimum of \( p_0 \)?

\begin{wrapfigure}{r}{0.16\textwidth}
    \centering
    \includegraphics[width=0.16\textwidth]{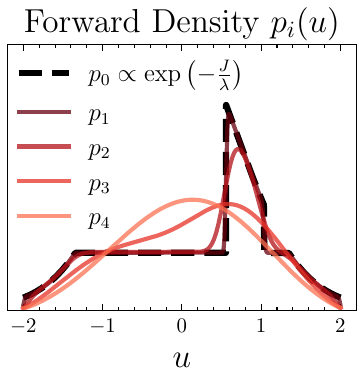}
    \caption{Forward density function in diffusion process.}
    \label{fig:demo_forward}
\end{wrapfigure}

\Cref{fig:demo_forward} illustrates that the convexity of the distribution increases with progressively larger kernels.  Each successive density function from $p_1$ to $p_4$ is obtained by convolving the original distribution $p_0$ with a progressively larger kernel. As discussed in \cref{subsec:sampling-mpc}, MPPI performs score ascent on \( p_i \). Therefore, the primary \textbf{advantage} of conducting score ascent on \( p_3 \) and \( p_4 \) is that MPPI can more easily converge to the global optimum.
The main \textbf{disadvantage} of optimizing a corrupted distribution is also clear from \cref{fig:demo_forward}. The optimum progressively shifts from $U^*$ to a distorted optimum under a larger kernel, thereby compromising the optimality of the original problem.

\noindent\textbf{Coverage and convergence trade-off.}
The advantages and disadvantages highlight a fundamental trade-off between exploration and exploitation in MPPI: a larger $\det{\Sigma}$ promotes greater exploration (i.e., coverage) by widening the sampling distribution, which helps to avoid suboptimal local minima.
However, a larger $\det{\Sigma}$ may compromise optimality(i.e., convergence) as it will introduce a larger optimality gap. In contrast, a smaller $\det{\Sigma}$ improves local optimality at the risk of getting trapped in local minima.
 
This trade-off becomes more pronounced in contact-rich tasks such as legged locomotion, where the optimization landscape is non-smooth, non-convex, and high-dimensional. 
The cost function $J$ and the distribution $p_0$ often have sharp and asymmetric peaks. 
In such cases, even a small increase in  $\det{\Sigma}$  can degrade performance, and global optimality cannot be guaranteed through score ascent on  $p_1(U)$  due to the small convolution kernel.

\begin{figure}[ht]
    \centering
    \includegraphics[width=\linewidth]{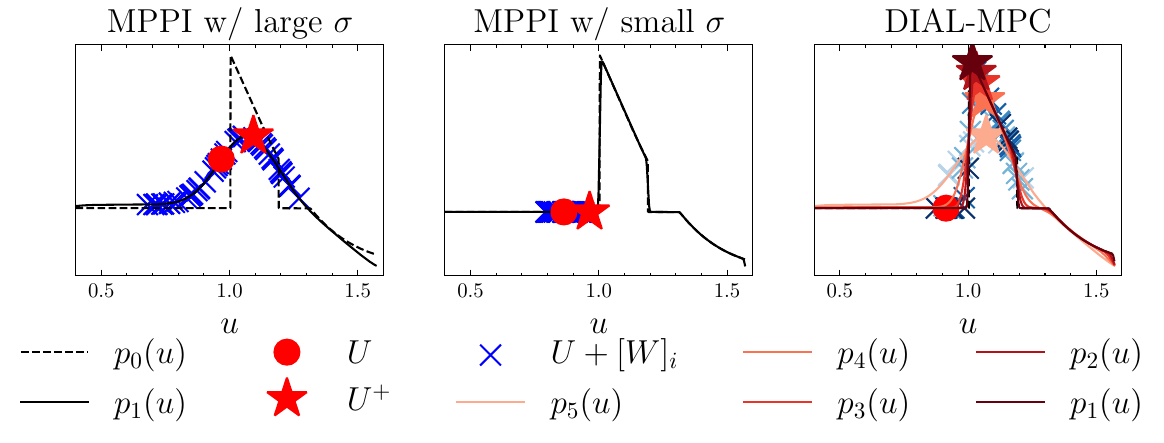}
    \caption{Coverage and convergence trade-off in sampling-based methods. Given the target distribution $p_0(U)$ and the same number of samples, MPPI either over-explores or over-exploits the solution, while our method balances the exploration and exploitation to converge to optima following a diffusion-inspired annealing process. }
    \label{fig:demo_compare}
\end{figure}

By fixing the sampling kernel size, MPPI may either over-explore or over-exploit, failing to balance exploration and convergence effectively. This concern is illustrated in \cref{fig:demo_compare}, which shows how different kernel sizes affect the performance of MPPI. Therefore, designing an effective sampling strategy that balances coverage and convergence is crucial to unlocking the potential of MPPI in real-world legged locomotion tasks.

Fortunately, diffusion processes, known for their powerful sampling capabilities from complex distributions, offer a way to balance coverage and convergence through an annealing strategy. This strategy involves sampling iteratively at decreasing noise levels, effectively reversing the forward corruption process. 

\noindent\textbf{Annealing in diffusion process.}
Instead of sampling solely from $p_1(\cdot)$, the forward process in diffusion defines a sequence of distributions with increasing noise levels: $ p_1(\cdot), \ldots, p_{N-1}(\cdot), p_N(\cdot)$, where $N$ is the total number of diffusion stages. Each density is defined as $p_i(\cdot) = (p_0 \ast \phi_i)(\cdot)$ with $\phi_i(\cdot) \sim \mathcal{N}(0, \Sigma^i)$. Sampling proceeds in reverse order. Starting from a higher noise level $\Sigma^N$, the sampling distribution $p_N(\cdot)$ is highly spread out, ensuring good global exploration.
As the noise level decreases, the sampling distributions  $p_i(U)$  become more concentrated, refining the search towards the target distribution  $p_0(U)$  and improving convergence to the optimal solution.
We adopt an exponential schedule for the noise levels, inspired by diffusion processes, to design the sampling kernels:
\begin{align}
\label{eq:exp_schedule}
    \det\left(\Sigma^{i}\right) = \exp\left(-\frac{N-i}{\beta N} d\right),\quad \forall i \in \{N, \ldots, 1\},
\end{align}
where $\beta$ is the temperature parameter for the annealing process, and $N$ is the number of iterations for the annealing process, $d$ is the dimension of the sampling space.

Given that MPPI corresponds to a single-stage diffusion process, we can naturally incorporate this multi-stage annealed diffusion approach to design an improved sampling strategy for MPPI. In~\cref{subsec:annealing_mppi}, we discuss how to implement this diffusion-inspired annealing process within the MPPI framework in a receding horizon manner.

\subsection{Diffusion-Inspired Annealing for Sampling-Based MPC}
\label{subsec:annealing_mppi}

\begin{algorithm}[h]
    \caption{Diffusion-Inspired Annealing for Legged MPC}
    \label{alg:asynchronous_annealing}
    \begin{algorithmic}[1]
        \STATE Initialize $u_{0:H} \leftarrow \mathbf{0}$
        \FOR{$t=0$ to $T$}
        \FOR{$i = 1$ to $N$}
        \STATE Get diffusion noise kernel $\Sigma^i_{t:t+H}$ with \eqref{eq:det_sync_async}.
        \STATE Sample $[W]_{1:N_W} \sim \mathcal{N}(0, \Sigma^{i}_{t:t+H})$.
        \STATE Rollout $u_{t:t+H} + [W]_{1:N_W}$ and evaluate the cost function $J(u_{t:t+H} + [W]_{1:N_W})$.
        \STATE Estimate Score $\nabla \log p_i(\cdot)$ with \eqref{eq:score}.
        \STATE Update $u_{t:t+H}^{(i)}$ with \eqref{eq:score_ascent}.
        \ENDFOR
        \STATE Receding horizon $u_{t+1:t+H+1} \leftarrow \texttt{shift}(u_{t:t+H})$
        \ENDFOR
    \end{algorithmic}
\end{algorithm}

Combining MPPI with multi-stage annealing, we propose \method (\cref{alg:asynchronous_annealing}) that leverages the receding horizon structure of MPC and introduces a dual-loop covariance design. 
This design comprises two annealing procedures: an outer-loop trajectory-level annealing and an inner-loop action-level annealing, which we detail below.

\noindent\textbf{Dual-loop annealing.}
In a receding horizon MPC framework with a horizon length $H$ and $N$ update iterations for each control sequence at each timestep $t$, $u_{H}$ (the last element of $u_{0:H}$) is first updated at time $0$ by $N$ times using convolution kernel $\Sigma^N_{H} \dots, \Sigma^1_{H}$. 
After $u_0$ is applied to the system, the control sequence shifts forward. 
At the next time step $t=1$, $u_{H}$ is updated again $N$ times, now with kernel $\Sigma^N_{H-1}, \dots, \Sigma^1_{H-1}$ as 
$u_H$ becomes the second-to-last element of the updated control sequence $u_{1:H+1}$. 
This procedure continues, and by the time  $u_H$  is applied to the system at  $t = H$, it will have undergone $N H$  updates with convolution kernels: $\Sigma^N_{H} \dots, \Sigma^1_{H} ; \Sigma^N_{H-1}, \dots, \Sigma^1_{H-1}; \dots; \Sigma^N_{0} \dots, \Sigma^1_{0}$ as shown in the last graph of \cref{fig:sync_async}.
Essentially, each control action  $u_h$ is updated in a dual-loop manner before being applied to the system. This motivates the design of two annealing procedures:
the outer loop is a trajectory-level annealing schedule for all the control sequence at a certain stage $i$ (i.e. designing the overall size of $\Sigma^i_{t:t+H}$ for a given $i$) and the inner loop is an action-level annealing schedule for different control at different horizons (i.e. the size of $\Sigma^i_{t+h}$ for $h=0,\ldots,H$).

\noindent\textbf{Trajectory-level annealing.}
For the outer-loop trajectory-level annealing, we define the schedule as:
\begin{align}
 \label{eq:det_sync}
\det\left(\Sigma^i_{t:t+H}\right) \propto \exp\left(-\frac{N-i}{\beta_1 N} H d_u \right), \quad \forall i \in \{N, \ldots, 1\},
\end{align}
where $\beta_1$ is the temperature parameter for the trajectory-level annealing, and $d_u$ is the dimension of a single control. The covariance matrix decreases over time as $i$ decreases, progressively narrowing the sampling distribution towards the target distribution $p_0$.

\noindent\textbf{Action-level annealing.}
For the inner-loop action-level annealing, the schedule is given by:
\begin{equation}
\label{eq:det_async}
    \det\left(\Sigma^i_{t+h}\right) \propto \exp\left(-\frac{H-h}{\beta_2 H} d_u \right), \quad \forall h \in \{0, \ldots, H\}, 
\end{equation}
where $\beta_2$ is the temperature parameter for action-level annealing. The covariance matrix increases with time as $h$ increases, 
allowing for a larger sampling region for future control actions that have been updated fewer times compared to those at the front of the horizon.

Note that \eqref{eq:det_sync} and \eqref{eq:det_async} specify only the overall size (i.e., the determinant) of the convolution kernels, leaving flexibility in designing the exact covariance matrices $\Sigma^i_{t+h}$. 
As a practical realization of~\eqref{eq:det_sync} and \eqref{eq:det_async}, we assume that each $\Sigma^i_{t+h}$ is isotropic and define it as:
\begin{align}
    \Sigma^i_{t+h} = \exp\left(-\frac{N-i}{\beta_1 N} - \frac{H-h}{\beta_2 H} \right) I.
    \label{eq:det_sync_async}
\end{align}

where $I$ is the identity matrix. This design combines both the outer-loop and inner-loop annealing schedules, adjusting the covariance matrices to ensure appropriate exploration and exploitation at each iteration and horizon step.

\section{EXPERIMENT}

In this section, we demonstrate the advantages of \method in terms of: (1) convergence and coverage, (2) efficiency in test-time generalizability, and (3) robustness to real-world model mismatch. 
We compare DIAL-MPC against MPPI and another sampling-based optimization method, CMA-ES~\cite{akimotoTheoreticalFoundationCMAES2012}. 
In addition, we provide goal-conditioned reinforcement learning (GCRL) as a performance reference. 
Our results show that \method reduces the tracking error by $3.9$ times in walking tasks compared to MPPI and outperforms GCRL on all tasks requiring both precision and agility, especially in the presence of significant model mismatch.

We design a set of agile locomotion tasks to showcase the performance of \method: (1) Walking Tracking: a quadruped robot\footnote{Detailed information of the hardware setup can be found in Appendix \ref{subsec:Hardware}.} is tasked with tracking a desired linear velocity and a desired yaw rate, requiring precise control of the torso. 
(2) Sequential Jumping: a quadruped robot must jump onto a series of small circular platforms placed randomly, each with a radius of \SI{10}{\centi\meter}. 
(3) Crate Climbing: a quadruped robot is tasked with climbing a crate with a height of \SI{60}{\centi\meter}, which is more than twice the height of the robot.  We leave the specific details of the tasks' implementation in Appendix \ref{subsec:task_imp}.

\subsection{Convergence and Coverage}

To answer the question of whether \method is a better solver for the legged MPC problems, we compare the performance of \method with vanilla MPPI, CMA-ES and NMPC.
Since vanilla MPPI only uses a single kernel size, we tested both MPPI with large kernel size $0.2$ (MPPI-explore) and small kernel $0.05$ (MPPI-exploit). 
Compared with \method, CMA-ES optimizes the sampling covariance in an evolutionary way.
For all sampling-based MPC methods, we use the same number of samples $N_W=2048$ and optimization steps $H=20$ (\SI{0.4}{\second}) to ensure a fair comparison. 
For NMPC, we use Mujoco MPC~\cite{howellPredictiveSamplingRealtime2022} as our baseline.
As a reference, we also include the performance of GCRL, which is trained using PPO~\cite{schulmanProximalPolicyOptimization2017} over 500 million steps, which takes approximately 31 minutes. 
We share the same reward functions for all methods. 
Due to unstable exploration in GCRL, we added two additional reward terms to regularize the policy, whereas \method only requires six reward terms.

\begin{table}[h]
    \centering
    \begin{tabular}{rccc}
        \toprule
                          & Walk Track $\downarrow$ & Seq Jump $\uparrow$ & Crate Climb $\uparrow$ \\ \midrule
        MPPI-explore      & 0.190                         & 0.440                  & 0.3                       \\
        MPPI-exploit      & 0.230                         & 0.450                  & 0.2                       \\
        CMA-ES            & 0.544                         & 0.345                  & 0.2                       \\
        NMPC              & 0.055                         & -                      & -                         \\
        GCRL$^*$             & 0.119                         & 0.855                  & 0.6                       \\
        \textbf{DIAL-MPC} & \textbf{0.024}                & \textbf{0.885}         & \textbf{0.9}              \\ \bottomrule
    \end{tabular}
    \caption{Performance comparison of different methods over three tasks. For walk tracking, it represents the tracking error; For sequential jumping, it represents average total contact reward; For crate climbing, it represents success rate out of 10 trials of climbing onto random crates with heights ranging from 0.125 to 0.6 \unit{\meter}. $^*$GCRL requires offline training so it is not an apple-to-apple baseline.}
    \label{tab:convergence}
\end{table}

\begin{figure}[h]
    \centering
    \includegraphics[width=1.0\linewidth]{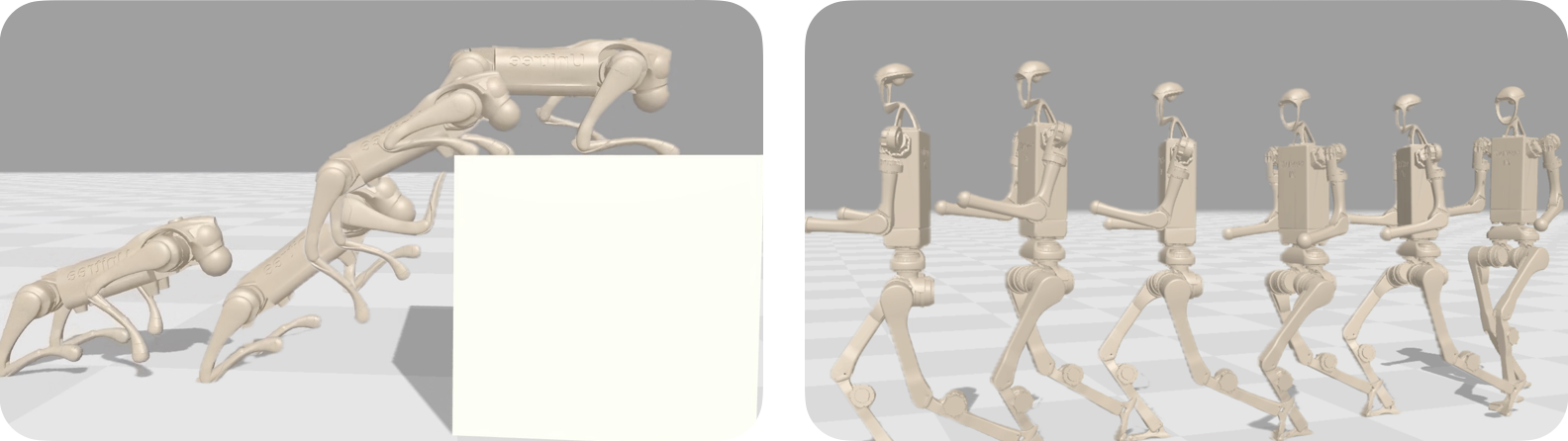} \\
    \vspace{0.15 cm}
    \includegraphics[width=1.0\linewidth]{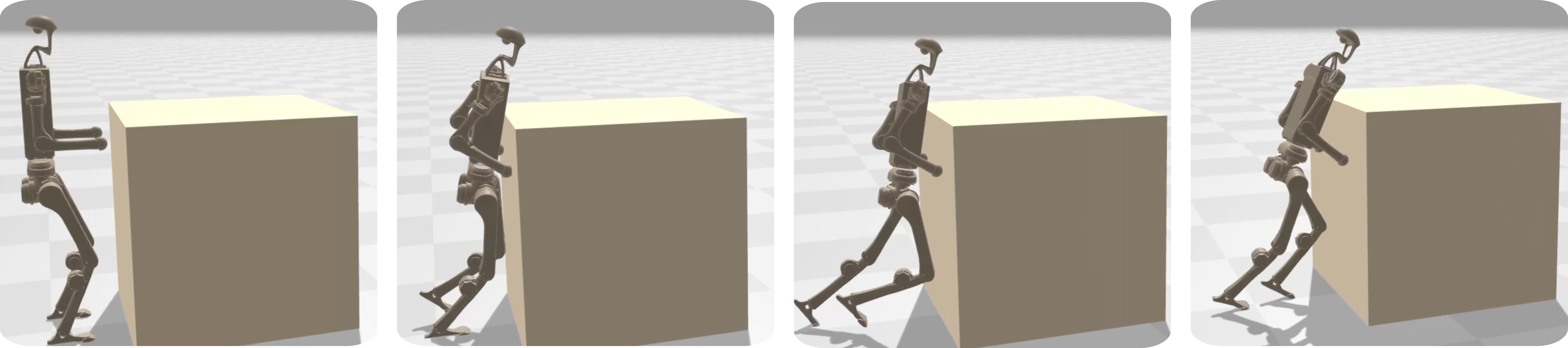} \\
    \vspace{0.15 cm}
    \includegraphics[width=1.0\linewidth]{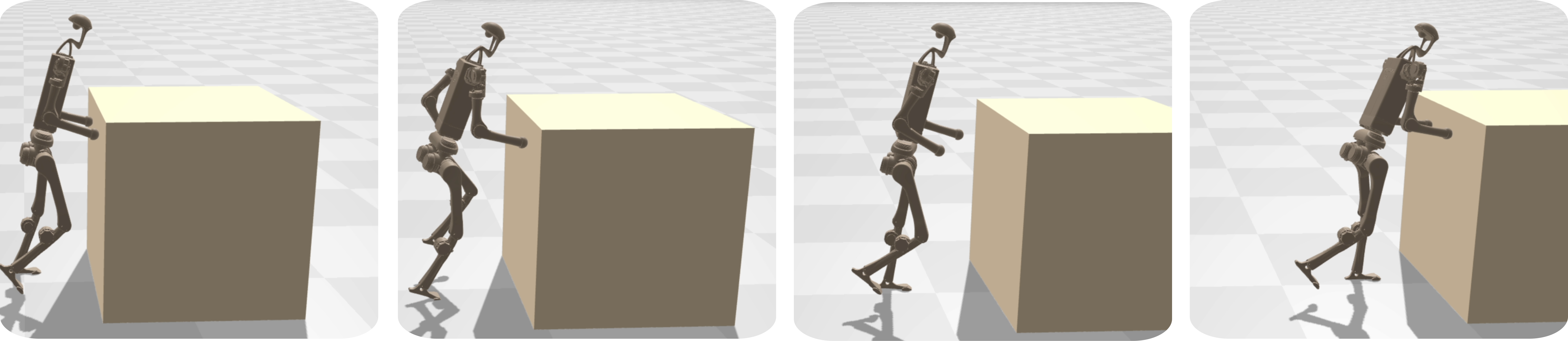}
    \caption{\textbf{Top}: the coverage of \method in crate-climbing and humanoid jogging task. The crate-climbing task requires the robot to climb up a crate more than two times higher than itself. The full-size humanoid jogging task demands DIAL-MPC to handle a higher-dimensional action space of 19. \textbf{Middle}: \method controlling a humanoid pushing a \SI{30}{\kilo\gram} crate. \textbf{Bottom}: \method generates a motion strategy with less effort when reducing the crate's weight to \SI{15}{\kilo\gram}.}
    \label{fig:coverage}
\end{figure}

\textbf{Convergence of \method.}
As shown in \cref{tab:convergence}, DIAL-MPC achieves the best performance across all tasks. 
Compared with other sampling-based MPC methods (namely MPPI and CMA-ES), DIAL-MPC generates better solutions, with $13.4$ times lower tracking error and $107.7\%$ higher contact reward, thanks to its diffusion-inspired annealing process. 
For the crate climbing task, DIAL-MPC is the only sampling-based MPC that consistently generates feasible solutions, improving the success rate by $3$ times compared to the best MPPI scheduling. 
This highlights the superior coverage of the solution space by \method. 
Compared with NMPC, \method is able to solve tasks requiring higher agility and non-smooth costs, such as sequential jumping and crate climbing, where NMPC is more likely to get stuck in local minima and fail to converge.

The superior convergence of \method is further demonstrated when compared with GCRL, where \method consistently outperforms GCRL in all tasks even if we use a lower-level leg position controller for RL and \method directly outputs the torque. 
Although DIAL-MPC is training-free, making a direct comparison potentially unfair, these results showcase the advantage of the annealing process in generating finer solutions compared to the Gaussian exploration in RL.

\textbf{Coverage of \method.}
DIAL-MPC is also capable of searching for non-trivial solutions and generating diverse motions. As examples, we visualize the solutions in the crate climbing task and a humanoid jogging task in \Cref{fig:coverage}. \method generates diverse solutions in high-dimensional spaces, thanks to the annealing process.

\subsection{Test-Time Generalizability }
As a training-free method, \method offers better test-time generalizability. Given a new task or model, DIAL-MPC can generate solutions in real time without finetuning. 

\begin{figure}
    \centering
    \includegraphics[width=1.0\linewidth, trim={0 0 1cm 0}]{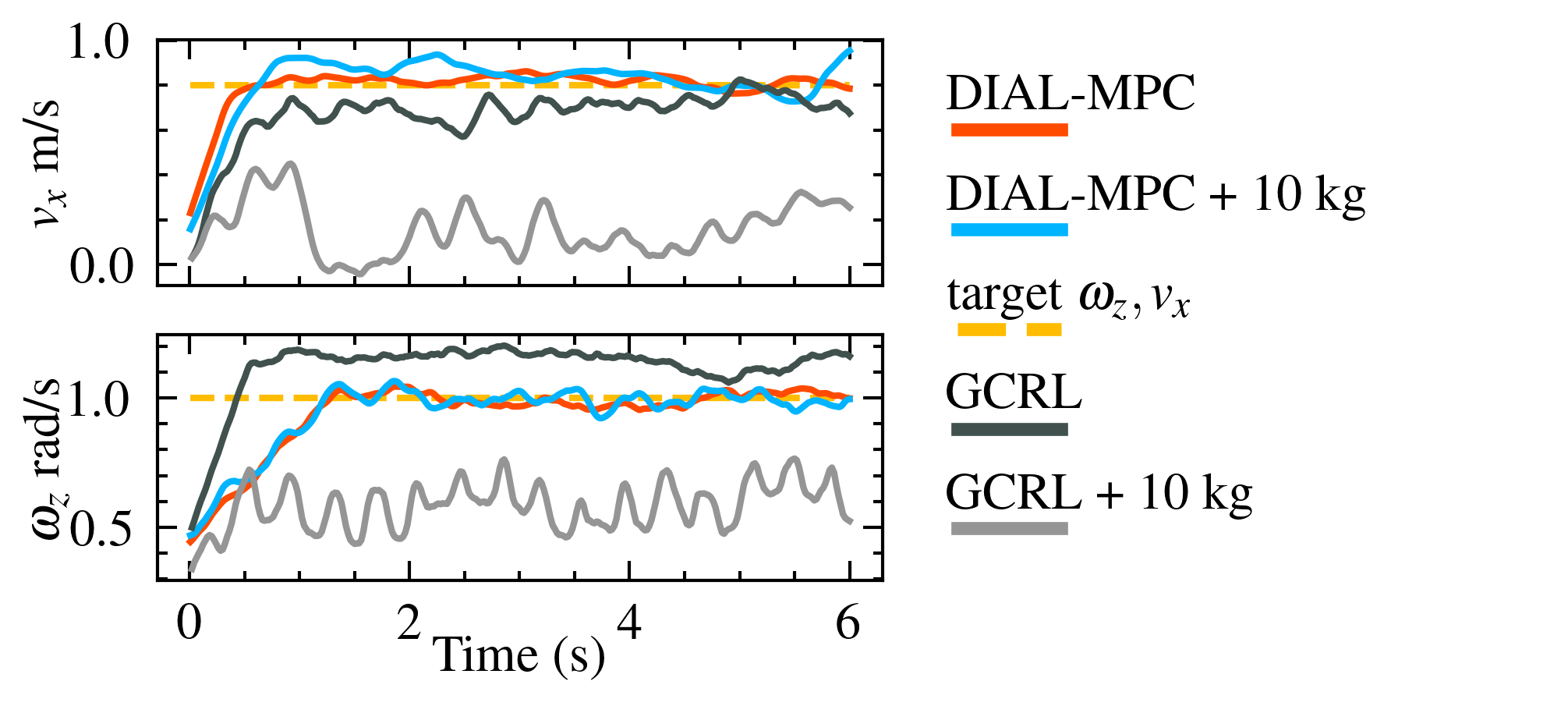}
    \caption{The linear and angular velocity tracking performance of \method and GCRL in the walking task test in simulation. GCRL fails with a $10$kg payload.}
    \label{fig:velocity-tracking}
    \vspace{-0.5cm}
\end{figure}

\textbf{Task-level generalizability. }
Compared with GCRL, DIAL-MPC reduces the tracking error by $3.9$ times in the walking task and improves the contact reward by $3.5\%$ given different goals.
\Cref{fig:velocity-tracking} visualizes the tracking performance of both methods. \method achieves higher tracking accuracy thanks to its explicit conditioning on each task, whereas GCRL uses a universal policy for all tasks.

\textbf{Dynamic-level generalizability. }
Another advantage of \method is its explicit conditioning on the dynamics model, enabling better adaptation to new dynamics given updated models. 
\Cref{fig:velocity-tracking} illustrates the tracking performance of both methods with a \SI{10}{\kilo\gram} payload attached to the robot’s base. The crate-climbing task is deprecated due to the heavy payload leads to infeasible solutions.
The GCRL policy is augmented with domain randomization of mass, actuator gain, and friction to improve robustness to model parameters. 
\Cref{tab:generalizability} shows the performance comparison of GCRL and \method under a \SI{10}{\kilo\gram} payload.
Without explicit conditioning on the physical parameters, GCRL performs poorly in tracking and completely fails after adding the payload. 
In contrast, \method outperform GCRL with a larger margin in the jumping task, demonstrating the advantage of explicitly conditioning on the dynamics model. 
While RL can be enhanced by conditioning on physical parameters or history, this requires additional training time and engineering effort. 
Conversely, the training-free DIAL-MPC can be instantly deployed with an updated model.

\begin{table}[h]
    \centering
    \begin{tabular}{rcc}
        \toprule
                          & Walk Track $\downarrow$ & Seq Jump $\uparrow$ \\ \midrule
        GCRL              &  0.517                             &   0.150                     \\
        \textbf{DIAL-MPC} &  \textbf{0.036}                             &   \textbf{0.815}                     \\ \bottomrule
    \end{tabular}
    \caption{Performance comparison of GCRL and \method under \SI{10}{\kilo\gram} payload.}
    \label{tab:generalizability}
\end{table}

\begin{figure}[h]
    \centering
    \includegraphics[width=1.0\linewidth]{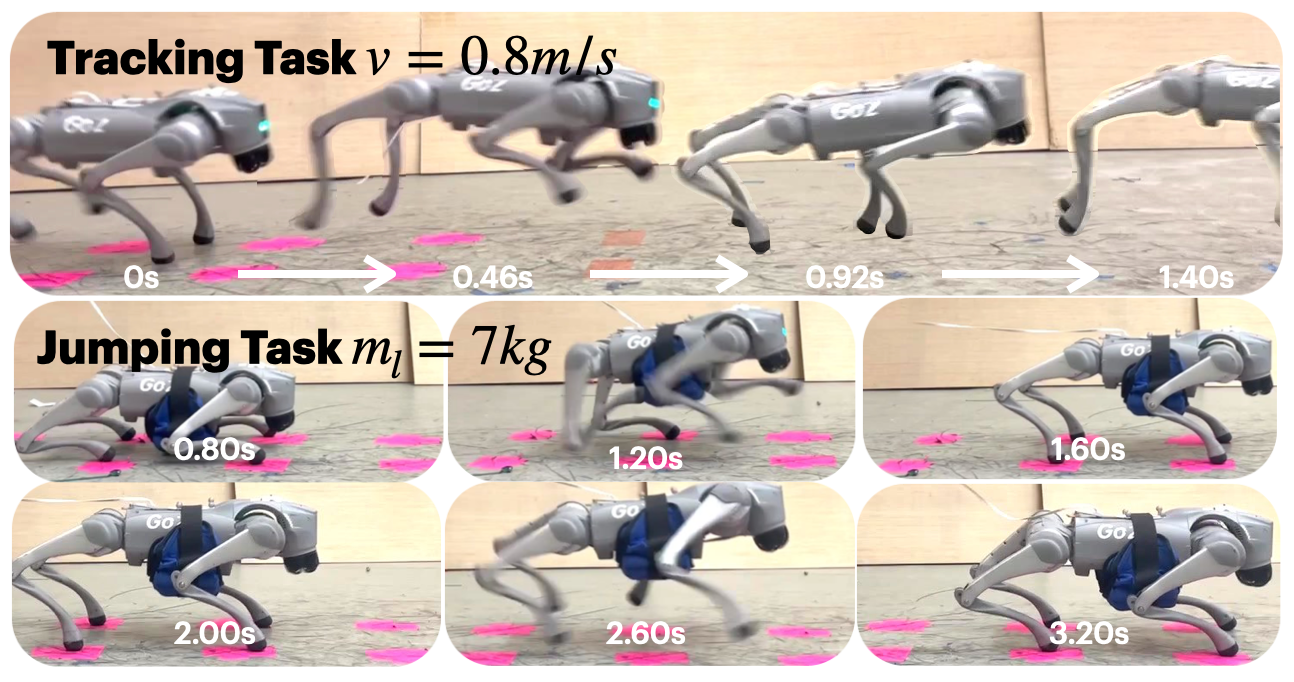}
    \caption{Velocity tracking task and sequential jumping task with a \SI{7}{\kilo\gram} payload on Go2 quadruped in real world with \method for direct torque control. \method enables both precise and agile motion.}
    \label{fig:real-world}
\end{figure}

\subsection{Robustness to Model Mismatch}

When dealing with unknown dynamics models, \method’s robustness stems from the noise injection process in the diffusion-inspired annealing. 
By controlling the final noise level, we can achieve a suboptimal but robust solution given a mismatched model. 
Since the sampling-based MPC baselines fail to operate effectively in the real world, we compare the performance of \method with the baseline in simulation with mismatched mass parameters.
\Cref{tab:real-world} shows the performance comparison of different methods under \SI{2}{\kilo\gram} base mass mismatch in simulation, where DIAL-MPC still outperforms the best sampling-based MPC baselines by $92\%$ in the walking task and $126\%$ in the jumping.

\begin{table}[h]
    \centering
    \begin{tabular}{rcc}
        \toprule
                          & Walk Track $\downarrow$ & Seq Jump $\uparrow$ \\ \midrule
        MPPI-explore      &  0.361                             & 0.270                       \\
        MPPI-exploit      &  0.407                             & 0.200                       \\
        CMA-ES            &  0.556                             & 0.100                       \\
        \textbf{DIAL-MPC} &  \textbf{0.188}                             & \textbf{0.610}                       \\ 
        \midrule
        DIAL-MPC in real &  0.322                             & 0.600                       \\
        \bottomrule
    \end{tabular}
    \caption{Performance comparison of different methods under \SI{2}{\kilo\gram} base mass mismatch in simulation and real world.}
    \label{tab:real-world}
\end{table}

\Cref{fig:real-world} shows the \method control a Unitree Go2 robot to walk and jump. The robot is able to walk smoothly and jump precisely with direct torque control.

\section{CONCLUSION}

This work presents \method, a sampling-based MPC method with a diffusion-inspired annealing process to balance coverage and convergence in real-world legged locomotion. \method can solve the full-order control problem efficiently with the help of diffusion process and is generalizable to various tasks and dynamics in a training-free manner. 
One limitation is that \method requires fast simulation to generate samples, which limits the application of \method in longer planning horizon tasks. 
In the future, we plan to further accelerate and improve the sample efficiency by learning a nominal policy, value function and model as what has been done in model-base RL.

\printbibliography

\appendix
\section{APPENDIX}

\subsection{Hardware and Software Setup}
We discuss various hardware platforms, including quadrupeds and humanoids, and provide details on the implementation of the algorithm.

\label{subsec:Hardware}
\noindent \textbf{Quadraped Configurations} For quadrupedal tasks, we use a Unitree GO2 robot with 18 degrees of freedom (DoF) and 12 actuated joints. The abduction and hip joints can produce \SI{24}{\newton\meter} torque, and the knee joints can produce \SI{45}{\newton\meter} torque approximately. We perform basic system identification to match the joint response to that in the simulation as closely as possible. The low-level controller for each joint of the quadruped plays a crucial role across various tasks. To ensure fair comparisons, the PID controller parameters are kept consistent across different testbeds. In the GCRL reference implementation, which uses position control, we set the proportional gain to 30 and the derivative gain to 0.65. In our method, which utilizes torque control, we apply the same derivative gain of 0.65 to regulate excessively high motor speeds. This choice also helps stabilize the simulation, which uses the same derivative gain. Given a torque command $\tau_i$, the final torque command sent to the $i$-th joint is
\begin{equation}
    \hat{\tau}_i = \tau_i - d\omega_i,
\end{equation}
where $i$ is the index of the joint, $d=0.65$ is the derivative gain and $\omega_i$ is the angular velocity of the $i$-th joint.

\noindent \textbf{Algorithm Implementation:} All algorithms are implemented in JAX~\cite{bradburyGoogleJax2024}, using the Brax simulator~\cite{freemanBraxDifferentiablePhysics2021}. The GCRL model is trained using the reference PPO implementation provided by Brax. All inference runs are performed on a desktop equipped with an RTX 4090 GPU, i9-13900KF CPU, and 32 GB of RAM, with the computer communicating with the robot via an Ethernet connection. For real-time control, all algorithms operate at \SI{50}{\hertz}, while a low-level controller repeats the last algorithm output and sends motor commands to the robot at \SI{200}{\hertz}. Ground-truth state estimation is obtained via a motion capture system. All sampling-based MPCs utilize 2048 parallel environments and a 20-step horizon (\SI{0.4}{\second}). To further speed up the simulations, contact forces are disabled for all robot parts except the feet.

\subsection{Task Implementation Details}
\label{subsec:task_imp}
\noindent\textbf{Quadruped Velocity Tracking}
We give all sampling-based MPCs the rewards specified in \cref{tab:walk-reward}. In particular, the gait is generated by a foot height generator outputting a trotting gait; energy is calculated as the product of joint torque $\tau_i$ and joint velocity $\omega_i$. 
With only six reward terms, \method demonstrates robust walking behavior in this task. We then applied GCRL using the same reward terms; to regularize the training process and improve the behavior, two additional key terms are introduced. The first is an alive reward to counterbalance the negative rewards and promote meaningful exploration during the initial stages of training. The second is a control rate cost, which subsumes the spline reparameterization trick used in \method to produce smoother actions. Furthermore, we implemented standard RL domain randomizations, varying the base mass by up to 1 kg, actuator gains by up to 20\%, and the friction coefficient between 0.6 and 1.0. For the goal, we randomized the linear velocity target to $\pm(1.5, 0.5, 0.0)~\unit{\meter\per\second}$, and the yaw angular velocity target to $\pm1.5~\unit{\radian\per\second}$. We trained GCRL using PPO for 500 million steps, which took 31 minutes.

\begin{table}[]
    \begin{tabular}{r|cc|cc}
    \toprule
    \multicolumn{1}{c|}{\textbf{}} & \multicolumn{2}{c|}{\textbf{Walking and Tracking}} & \multicolumn{2}{c}{\textbf{Sequential Jumping}} \\ \midrule
    \textbf{} & \textbf{\method} & \textbf{GCRL} & \textbf{\method} & \textbf{GCRL} \\ \midrule
    \textbf{Upright} & -1.0 & -1.0 & -1.0 & -1.0 \\
    \textbf{Base Height} & -1.0 & -1.0 & -1.0 & -1.0 \\
    \textbf{Energy} & -0.001 & -0.001 & -0.001 & -0.001 \\
    \textbf{Linear Velocity} & -0.5 & -1.0 & - & - \\
    \textbf{Angular Velocity} & -0.5 & -1.0 & - & - \\
    \textbf{Gait} & -0.1 & -0.1 & \textbf{-} & \textbf{-} \\
    \textbf{Contact Reward} & - & - & 0.1 & 0.1 \\
    \textbf{Contact Penalty} & - & - & -0.1 & -0.1 \\
    \textbf{Alive} & - & 3.0 & - & 10.0 \\
    \textbf{Control Rate} & - & 0.001 & - & 0.001 \\ \bottomrule
    \end{tabular}
    \caption{Reward specifications for the quadruped walking-tracking and sequential jumping tasks are largely shared between our method and the GCRL baseline, underscoring the flexibility of \method in utilizing RL-style rewards.}
    \label{tab:walk-reward}
    \end{table}

\begin{figure}
    \centering
    \includegraphics[width=1.0\linewidth]{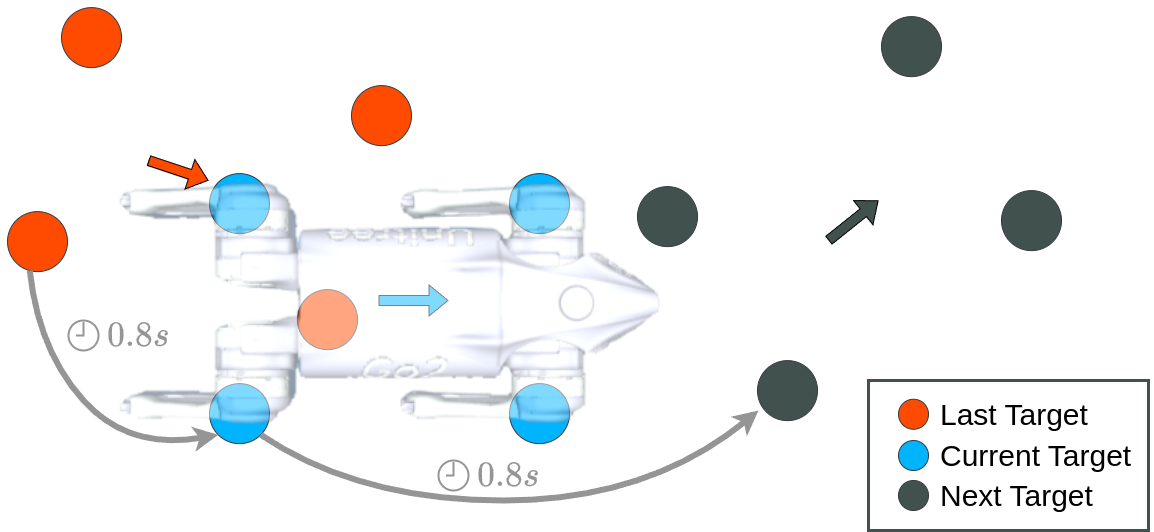}
    \caption{A visual illustration of the sequential jumping task. At every stage, the quadruped is given \SI{1}{\second} to jump onto the next target. Contacts outside the target are penalized.}
    \label{fig:jumping-task-illustration}
\end{figure}

\noindent\textbf{Quadruped Sequential Jumping}
In this task, quadruped must rapidly jump onto a series of \SI{10}{\centi\meter} plates. 
The task is depicted in Figure \ref{fig:jumping-task-illustration}. At a regular time interval, a new contact target is given in terms of four-foot placements and a center-of-mass (CoM) location. The interval is so short (\SI{1}{\second}) that the quadruped must transition smoothly and dynamically between the targets. We uniformly sample the next contact target, whose CoM location is within $\pm(0.65, 0.65, 0.0)$\unit{\meter} translation and \SI{0.5}{\radian} yaw rotation w.r.t. the current contact target.

Inspired by WoCoCo \cite{zhangWoCoCoLearningWholeBody}, we use a staged contact reward system at time $t$:
\begin{equation}
    r_{\text{con}}^{(j)}(t) = w_{\text{correct}}n_{\text{correct}}^{(j)}(t)  - w_{\text{wrong}}\left(n_{\text{wrong}}^{(j)}(t) - n_{\text{correct}}^{(j-1)}(t)\right),
\end{equation}
where $r_{\text{con}}^{(j)}$ is the current-stage contact reward function; $w_{\text{correct}}$ and $w_{\text{wrong}}$ are the contact reward and penalty weights; $n_{\text{correct}}^{(j)}$ and $n_{\text{wrong}}^{(j)}$ compute the correct and wrong number of contacts w.r.t. the current stage target; $n_{\text{correct}}^{(j-1)}$ computes the correct number of contacts w.r.t. the last stage target; The last term offsets the penalty of wrong contacts that were valid last-stage contacts. This prevents the robot from being penalized or rewarded while preparing for its current jump.

We establish a performance measurement using the total contact reward over a sequence of 10 jumping stages in \SI{8}{\second}, denoted as $R_{\text{con}}$:
\begin{equation}
    R_{\text{con}} = \sum_{j=1}^{j_{\text{max}}} \min_{t \in [T_{\text{min}}(j), T_{\text{max}}(j))}\left(r_{\text{con}}^{(j)}(t)\right),
\end{equation}
where $j_{\text{max}}=10$; $T_{\text{min}}(j), T_{\text{max}}(j)$ maps the time interval of the $j$-th contact stage; the minimum function finds the worst-case contact score at every stage.

We let every algorithm perform 5 trials, and report the average total contact reward, $\bar{R}_{\text{con}}$. Same as in the walking-tracking experiment, we add a \SI{10}{\kilo\gram} payload to the robot, and update the model parameters in all sampling-based MPCs accordingly.

\begin{figure}[h]
    \centering
    \includegraphics[width=\linewidth]{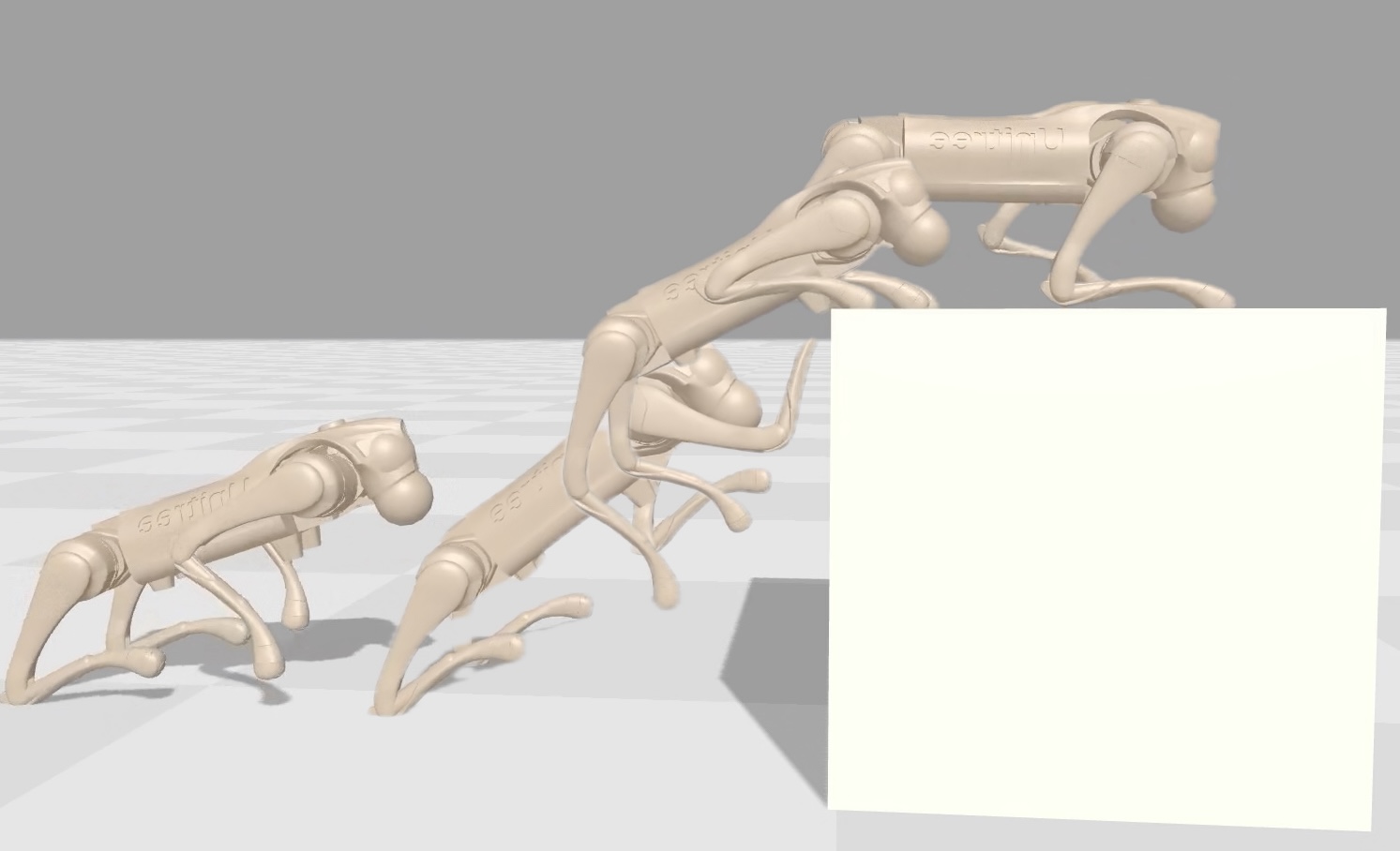}
    \caption{\method generates this well-planned agile motion in 30 seconds, training free. The quadruped leaps forward to catch the edge of the crate with its forelegs, launches off the ground with its hindlegs, and quickly retracts to a standing pose on the crate.}
    \label{fig:crate-jumping}
\end{figure}

\noindent\textbf{Quadruped Crate Climbing}
In \cref{fig:crate-jumping}, we demonstrate \method's ability to achieve RL-level agility in 1/15 the speed of real world. The quadruped is tasked with climbing onto the \SI{0.6}{\meter} high crate, which is more than twice the standing height of the robot. Similar tasks have been demonstrated on several recent quadruped parkour works with RL \cite{chengExtremeParkourLegged2024}, which report around 20 hours of training time for a single policy with perception. Although \method's integration with perception demands future work, it is currently comparable to a teacher policy with access to privileged environment information.

We use only five reward terms to guide \method to complete the task. They are shown in Table \ref{tab:crate-climbing-reward}. Contact reward is given to feet on the top surface of the crate. Since this is a non-real-time planning task, we use 4096 parallel samples, 40-step (\SI{1}{\second}) horizon, and 4 annealing steps. We also enable all contacts on the base and thighs. We then roll out the simulation synchronously after \method finishes computing at every step, which in total takes 30 seconds to generate a 2-second full-state motion plan with zero prior knowledge or training.

\begin{table}[]
\centering
    \begin{tabular}{rc}
        \toprule
                                 & \textbf{Crate Climbing} \\ \midrule
        \textbf{Target Position} & 0.5                     \\
        \textbf{Upright}         & 0.01                    \\
        \textbf{Energy}          & 0.001                   \\
        \textbf{Target Yaw}      & 0.3                     \\
        \textbf{Contact Reward}  & 0.2                     \\ \bottomrule
    \end{tabular}
    \caption{Reward weights of the crate climbing task.}
    \label{tab:crate-climbing-reward}
\end{table}

\noindent\textbf{Humanoid Crate Pushing}

Figure \ref{fig:coverage} shows \method performing whole-body control task on a Unitree H1 humanoid in simulation. The robot is asked to track a slow velocity of \SI{0.8}{\meter\per\second} while pushing forward a crate of \SI{30}{\kilo\gram} and \SI{15}{\kilo\gram} respectively. The robot is rewarded for maintaining the correct speed and the correct contact with its two upper end effectors. The seven reward weights are listed in Table \ref{tab:crate-pushing-reward}. This task shows that \method generalizes well to systems with higher degrees of freedom (25 for the humanoid and 18 for the quadruped) and is capable of handling whole-body control tasks with versatile parameters. Again we highlight that all reward terms in this experiment are straight forward, and that it takes no time to visualize the outcome of reward tuning since \method is a training-free algorithm.

\begin{table}[]
\centering
    \begin{tabular}{rc}
        \toprule
                                 & \textbf{Humanoid Crate Pushing} \\ \midrule
        \textbf{Gait} & 5.0                     \\
        \textbf{Upright}         & 0.01                    \\
        \textbf{Yaw}         & 0.1                    \\
        \textbf{Velocity}         & 1.0                    \\
        \textbf{Torso Height}         & 0.5                    \\
        \textbf{Energy}          & 0.01                   \\
        \textbf{Contact Reward}  & 0.05                   \\ \bottomrule
    \end{tabular}
    \caption{Reward weights of the humanoid crate pushing task.}
    \label{tab:crate-pushing-reward}
\end{table}

\end{document}